\documentclass{article}
\usepackage{times}
\usepackage{enumerate}
\usepackage{amsthm}
\usepackage{amsmath}
\usepackage{amsfonts}
\usepackage{amssymb}
\usepackage{array}
\usepackage{latexsym}
\usepackage{coling08}
\usepackage{algorithm}
\usepackage{algpseudocode}
\usepackage{color}
\usepackage{graphics}
\usepackage{graphicx}
\title{How Is Meaning Grounded in Dictionary Definitions?}

\author{A. Blondin Mass\'e\\
	Laboratoire de combinatoire et d'informatique math\'ematique\\
	Universit\'e du Qu\'ebec \`a Montr\'eal\\
	Montr\'eal (QC), CANADA H3C 3P8\\
	{\tt alexandre.blondin.masse@gmail.com}
	\AND
	G. Chicoisne, Y. Gargouri, S. Harnad, O. Picard\\
	Institut des sciences cognitives\\
	Universit\'e du Qu\'ebec \`a Montr\'eal\\
	Montr\'eal (QC), CANADA H3C 3P8\\
	{\tt chicoisne.guillaume@uqam.ca, yassinegargouri@hotmail.com}\\
	{\tt harnad@ecs.soton.ac.uk, olivierpicard18@hotmail.com}
	\AND
	O. Marcotte\\
	Groupe d'\'etudes et de recherche en analyse des d\'ecisions (GERAD)
	and UQ\`AM\\
	HEC Montr\'eal\\
	Montr\'eal (Qu\'ebec) Canada H3T 2A7\\
	{\tt Odile.Marcotte@gerad.ca}
}

\date{}

\newtheorem{definition}{Definition}
\newtheorem{lemma}[definition]{Lemma}
\newtheorem{theorem}[definition]{Theorem}
\newtheorem{proposition}[definition]{Proposition}
\newtheorem{corollary}[definition]{Corollary}
\newtheorem{example}[definition]{Example}

\newenvironment{basis}{\vskip 0.5\baselineskip \noindent \textsc{Basis.}}
{}
\newenvironment{induction}{\vskip 0.5\baselineskip \noindent 
\textsc{Induction.}}{}

\newcommand{\kGS}{\mbox{$k$-GS}}
\newcommand{\kFVS}{\mbox{$k$-FVS}}
\newcommand{\NPcomplete}{\mbox{NP-complete}}
\newcommand{\et}{\, \& \,}

\newcommand{\ReachableBFS}{\textsc{ReachableSet}}
\newcommand{\GroundingKernel}{\textsc{GroundingKernel}}

\begin{document}
\maketitle
\begin{abstract}
Meaning cannot be based on dictionary definitions all the way down: at some 
point the circularity of definitions must be broken in some way, by grounding 
the meanings of certain words in sensorimotor categories learned from 
experience or shaped by evolution. This is the ``symbol grounding problem". We 
introduce the concept of a \emph{reachable} set --- a larger vocabulary whose 
meanings can be learned from a smaller vocabulary through definition alone, as 
long as the meanings of the smaller vocabulary are themselves already grounded. 
We provide simple algorithms to compute reachable sets for any given 
dictionary.
\end{abstract}

\section{Introduction}

We know from the 19th century philosopher-mathematician Frege that the 
\emph{referent} and the \emph{meaning} (or ``sense'') of a word (or phrase) 
are not the same thing: two different words or phrases can refer to the very 
same object without having the same meaning \cite{Freg48}: ``George W. Bush" 
and ``the current president of the United States of America" have the same 
referent but a different meaning. So do ``human females" and ``daughters". 
And ``things that are bigger than a breadbox" and ``things that are not the 
size of a breadbox or smaller".

A word's ``extension" is the set of things to which it refers, and its 
``intension" is the rule for defining what things fall within its extension.. A 
word's meaning is hence something closer to \emph{a rule for picking out its 
referent.} Is the dictionary definition of a word, then, its meaning?

Clearly, if we do not know the meaning of a word, we look up its definition in 
a dictionary. But what if we do not know the meaning of any of the words in its 
dictionary definition? And what if we don't know the meanings of the words in 
the definitions of the words defining those words, and so on? This is a problem 
of infinite regress, called the ``symbol grounding problem" 
\cite{Harn90,Harn03}: the meanings of words in dictionary definitions are, in 
and of themselves, ungrounded. The meanings of some of the words, at least, 
have to be grounded by some means other than dictionary definition look-up.

How are word meanings grounded? Almost certainly in the sensorimotor capacity 
to pick out their referents \cite{Harn05}. Knowing \emph{what to do with what} 
is not a matter of definition but of adaptive sensorimotor interaction between 
autonomous, behaving systems and categories of ``objects" (including 
individuals, kinds, events, actions, traits and states). Our embodied 
sensorimotor systems can also be described as applying information processing 
rules to inputs in order to generate the right outputs, just as a thermostat 
defending a temperature of 20 degrees can be. But this dynamic process is in no 
useful way analogous to looking up a definition in a dictionary.

We will not be discussing sensorimotor grounding \cite{Bars08,Glen02,Stee07} in 
this paper. We will assume some sort of grounding as given: when we consult a 
dictionary, we already know the meanings of at least some words, somehow. A 
natural first hypothesis is that the grounding words ought to be more concrete, 
referring to things that are closer to our overt sensorimotor experience, and 
learned earlier, but that remains to be tested \cite{Clar03}. Apart from the 
question of the boundary conditions of grounding, however, there are basic 
questions to be asked about the structure of word meanings in dictionary 
definition space. 

In the path from a word, to the definition of that word, to the definition of 
the words in the definition of that word, and so on, through what sort of a 
structure are we navigating \cite{Rava03,Stey05}? Meaning is compositional: A 
definition is composed of words, combined according to syntactic rules to form 
a proposition (with a truth value: true or false). For example, the word to be 
defined $w$ (the ``definiendum") might mean $w_1 \et w_2 \et \ldots \et w_n$, 
where the $w_i$ are other words (the ``definientes") in its definition. Rarely 
does that proposition provide the full necessary and sufficient conditions for 
identifying the referent of the word, $w$, but the approximation must at least 
be close enough to allow most people, armed with the definition, to understand 
and use the defined word most of the time, possibly after looking up a few of 
its definientes $d_w$, but without having to cycle through the entire 
dictionary, and without falling into circularity or infinite regress. 

If enough of the definientes are grounded, then there is no problem of infinite 
regress. But we can still ask the question: What is the size of the grounding 
vocabulary? and what words does it contain? What is the length and shape of the 
path that would be taken in a recursive definitional search, from a word, to 
its definition, to the definition of the words in its definition, and so on? 
Would it eventually cycle through the entire dictionary? Or would there be 
disjoint subsets? 

This paper raises more questions than it answers, but it develops the formal 
groundwork for a new means of finding the answers to questions about how word 
meaning is \emph{explicitly} represented in real dictionaries --- and perhaps 
also about how it is \emph{implicitly} represented in the ``mental lexicon" 
that each of us has in our brain \cite{Hauk08}.

The remainder of this paper is organized as follows: In Section 
\ref{S:definitions}, we introduce the graph-theoretical definitions and 
notations used for formulating the symbol grounding problem in Section 
\ref{S:formulation}. Sections \ref{S:psycholinguistics} and \ref{S:kernels} 
deal with the implication of this approach in cognitive sciences and show in 
what ways grounding kernels may be useful.

\section{Definitions and Notations} \label{S:definitions}

In this section, we give mathematical definitions for the dictionary-related 
terminology, relate them to natural language dictionaries and supply the 
pertinent graph theoretical definitions. Additional details are given to ensure 
mutual comprehensibility to specialists in the three disciplines involved 
(mathematics, linguistics and psychology). Complete introductions to graph 
theory and discrete mathematics are provided in \cite{Bond78,Rose07}.

\subsection{Relations and Functions}

Let $A$ be any set. A \emph{binary relation on $A$} is any subset $R$ of 
$A \times A$. We write $xRy$ if $(x,y) \in R$. The relation $R$ is said to be 
(1) \emph{reflexive} if for all $x \in A$, we have $xRx$, (2) \emph{symmetric} 
if for all $x,y \in A$ such that $xRy$, we have $yRx$ and (3) \emph{transitive} 
if for all $x,y,z \in A$ such that $xRy$ and $yRz$, we have $xRz$. The 
relation $R$ is an \emph{equivalence relation} if it is reflexive, symmetric 
and transitive. For any $x \in A$, the \emph{equivalence class of $x$}, 
designated by $[x]$, is given by $[x] = \{y \in A \mid xRy\}$. It is easy to 
show that $[x] = [y]$ if and only if $xRy$ and that the set of all equivalence 
classes forms a partition of $A$.

Let $A$ be any set, $f: A \rightarrow A$ a function and $k$ a positive 
integer. We designate by $f^k$ the function $f \circ f \circ \ldots \circ f$ 
($k$ times), where $\circ$ denotes the \emph{composition of functions}.

\subsection{Dictionaries}

At its most basic level, a dictionary is a set of associated pairs: a 
\emph{word} and its \emph{definition}, along with some disambiguating 
parameters. The \emph{word}\footnote{In the context of this mathematical 
analysis, we will use ``word" to mean a finite string of uninterrupted letters 
having some associated meaning.} to be defined, $w$, is called the 
\emph{definiendum} (plural: \emph{definienda}) while the finite nonempty set of 
words that defines $w$, $d_w$, is called the set of \emph{definientes} of $w$ 
(singular: \emph{definiens}).

Each dictionary entry accordingly consists of a definiendum $w$ followed by its 
set of definientes $d_w$. A \emph{dictionary} $D$ then consists of a finite set 
of pairs $(w, d_w)$ where $w$ is a word and $d_w = \{w_1,w_2,\ldots,w_n\}$, 
where $n \geq 1$, is its definition, satisfying the property that for all 
$(w, d_w) \in D$ and for all $d \in d_w$, there exists $(w', d_{w'}) \in D$ 
such that $d = w'$. A pair $(w, d_w)$ is called an \emph{entry} of $D$. In 
other words, a dictionary is a finite set of words, each of which is defined, 
and each of its defining words is likewise defined somewhere in the dictionary.

\subsection{Graphs} \label{SS:graphs}

A \emph{directed graph} is a pair $G = (V,E)$ such that $V$ is a finite set of 
\emph{vertices} and $E \subseteq V \times V$ is a finite set of \emph{arcs}. 
Given $V' \subseteq V$, the \emph{subgraph induced by $V'$}, designated by 
$G[V']$, is the graph $G[V'] = (V',E')$ where $E' = E \cap (V' \times V')$. For 
any $v \in V$, $N^-(v)$ and $N^+(v)$ designate, respectively, the set of 
incoming and outgoing neighbors of $v$, i.e.
\begin{eqnarray*}
	N^-(v) & = & \{u \in V \mid (u,v) \in E\} \\
	N^+(v) & = & \{u \in V \mid (v,u) \in E\}.
\end{eqnarray*}
We write $\deg^-(v) = |N^-(v)|$ and $\deg^+(v) = |N^+(v)|$, respectively. A 
\emph{path} of $G$ is a sequence $(v_1,v_2,\ldots,v_n)$, where $n$ is a 
positive integer, $v_i \in V$ for $i=1,2,\ldots,n$ and $(v_i,v_{i+1}) \in E$, 
for $i=1,2,\ldots,n-1$. A \emph{$uv$-path} is a path starting with $u$ and 
ending with $v$. Finally, we say that a $uv$-path is a \emph{cycle} if $u = v$.

Given a directed graph $G = (V,E)$ and $u,v \in V$, we write $u \rightarrow v$ 
if there exists a $uv$-path in $G$. We define a relation $\sim$ as
$$u \sim v \Leftrightarrow u \rightarrow v \mbox{ and } v \rightarrow u.$$
It is an easy exercise to show that $\sim$ is an equivalence relation. The 
equivalence classes of $V$ with respect to $\sim$ are called the 
\emph{strongly connected components} of $G$. In other words, in a directed 
graph, it might be possible to go directly from point $A$ to point $B$, without 
being able to get back from point $B$ to point $A$ (as in a city with only 
one-way streets). Strongly connected components, however, are subgraphs in 
which whenever it is possible to go from point $A$ to point $B$, it is also 
possible to come back from point $B$ to point $A$ (the way back may be 
different).

There is a very natural way of representing definitional relations using graph 
theory, thus providing a formal tool for analyzing grounding properties of 
dictionaries: words can be represented as vertices, with arcs representing 
definitional relations, i.e. there is an arc $(u,v)$ between two words $u$ and 
$v$ if the word $u$ appears in the definition of the word $v$. More formally, 
for every dictionary $D$, its \emph{associated graph} $G = (V,E)$ is given by
\begin{eqnarray*}
	V & = & \{w \mid \exists d_w \mbox{ such that } (w,d_w) \in D\}, \\
	E & = & \{(v,w) \mid \exists d_w \mbox{ such that } (w,d_w) \in D 
	                                                    \mbox{ and } \\
	  &   & v \in d_w\}.
\end{eqnarray*}
Note that every vertex $v$ of $G$ satisfies $\deg_G^-(v) > 0$, but it is 
possible to have $\deg_G^+(v) = 0$. In other words, whereas every word has a 
definition, some words are not used in any definition.

\begin{figure}[ht]
	\includegraphics[width=\columnwidth]{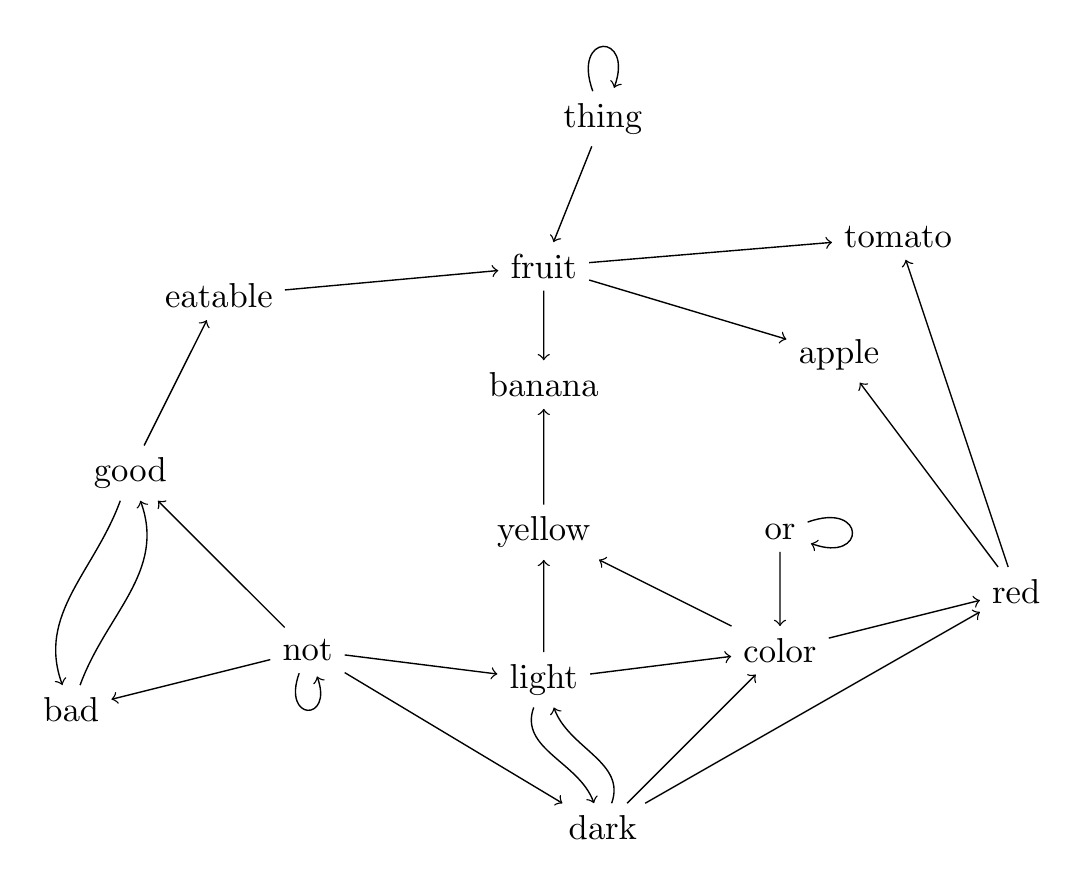}
	\caption{Graph representation of the dictionary $D$.}
	\label{F:graph-dictionary}
\end{figure}

\begin{example} \label{E:dictionary}
	Let $D$ be the dictionary whose definitions are given in Table 
	\ref{T:definitions}.	Note that every word appearing in some 
	definition is likewise defined in $D$ (this is one of the criteria for 
	$D$ to be a dictionary). The associated graph $G$ of $D$ is represented 
	in Figure \ref{F:graph-dictionary}. Note that $($not, good, eatable, 
	fruit$)$ is a path of $G$ while $($good, bad, good$)$ is a cycle (as 
	well as a path) of $G$.
\end{example}

\begin{table}[ht]
	\centering
	\begin{tabular}{|ll|ll|}
		\hline
		Word   & Definition     & Word    & Definition \\
		\hline 
		apple  & red fruit      & bad     & not good \\
		banana & yellow fruit   & color   & dark or light \\
		dark   & not light      & eatable & good \\
		fruit  & eatable thing  & good    & not bad \\
		light  & not dark       & not     & not \\
		or     & or             & red     & dark color \\
		thing  & thing          & tomato  & red fruit \\
		yellow & light color    &         & \\
		\hline
	\end{tabular}
	\caption{Definitions of the dictionary $D$}
	\label{T:definitions}
\end{table}

\section{A Graph-Theoretical Formulation of the Problem} \label{S:formulation}

We are now ready to formulate the symbol grounding problem from a mathematical 
point of view.

\subsection{Reachable and Grounding Sets}

Given a dictionary $D$ of $n$ words and a person $x$ who knows $m$ out of these 
$n$ words, assume that the only way $x$ can learn new words is by consulting 
the dictionary definitions. Can all $n$ words be learned by $x$ through 
dictionary look-up alone? If not, then exactly what subset of words can be 
learned by $x$ through dictionary look-up alone? 

For this purpose, let $G = (V,E)$ be a directed graph and consider the 
following application, where $2^V$ denotes the collection of all subsets of 
$V$:
\begin{center} $\begin{array}{lrcl}
R_G: & 2^V & \longmapsto & 2^V \\
     & U   & \longmapsto & U \cup \{v \in V \mid N^-(v) \subseteq U\}.
\end{array}$ \end{center}
When the context is clear, we omit the subscript $G$. Also we let $R^k$ denote
the $k^{\mbox{\footnotesize th}}$ power of $R$. We say that $v \in V$ is 
\emph{$k$-reachable from} $U$ if $v \in R^k(U)$ and $k$ is a nonnegative 
integer. It is easy to show that there exists an integer $k$ such that 
$R^\ell(U) = R^k(U)$, for every integer $\ell > k$. More precisely, we have the 
following definitions:

\begin{definition}
	Let $G = (V,E)$ be a directed graph, $U$ a subset of $V$, and $k$ an 
	integer such that $R^\ell(U) = R^k(U)$ for all $\ell > k$. The set 
	$R^k(U)$ is called the \emph{reachable set from $U$} and is denoted by 
	$R^*(U)$. Moreover, if $R^*(U) = V$, then we say that $U$ is a 
	\emph{grounding set} of $G$.
\end{definition}
We say that $G$ is \emph{$p$-groundable} if there exists $U \subseteq V$ such 
that $|U| = p$ and $U$ is a grounding set of $G$. The \emph{grounding number} 
of a graph $G$ is the smallest integer $p$ such that $G$ is $p$-groundable.

Reachable sets can be computed very simply using a breadth-first-search type 
algorithm, as shown by Algorithm \ref{A:reachableBFS}.

\begin{algorithm}[ht]
\caption{Computing reachable sets}
\label{A:reachableBFS}
\begin{algorithmic}[1]
	\Function{\ReachableBFS}{$G, U$}
		\State $R \leftarrow U$
		\Repeat
			\State $S \leftarrow \{v \in V \mid N^-_G(v) 
			                                \subseteq R\} - R$
			\State $R \leftarrow R \cup S$
		\Until{$S = \emptyset$}
		\State \Return $R$
	\EndFunction
\end{algorithmic}
\end{algorithm}

We now present some examples of reachable sets and grounding sets.

\begin{figure}[ht]
	\includegraphics[width=\columnwidth]{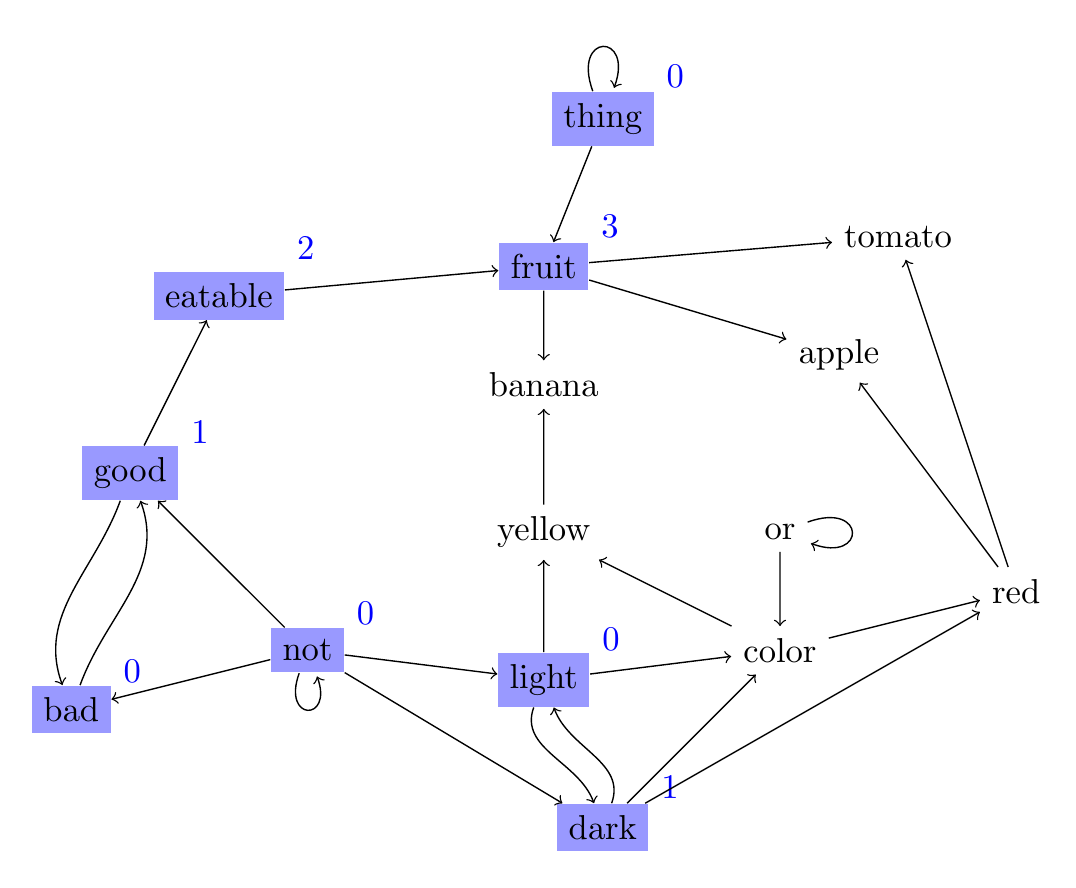}
	\caption{The set $R^*(U)$ (the words in squares) obtained from $U$}
	\label{F:reachable}	
\end{figure}

\begin{example}
	Consider the dictionary $D$ and the graph $G$ of Example 
	\ref{E:dictionary}. Let $U = \{$bad, light, not, thing$\}$. Note that
	\begin{eqnarray*}
		R^0(U) & = & U \\
		R^1(U) & = & U \cup \{\mbox{dark}, \mbox{good}\}, \\
		R^2(U) & = & R^1(U) \cup \{\mbox{eatable}\} \\
		R^3(U) & = & R^2(U) \cup \{\mbox{fruit}\} \\
		R^4(U) & = & R^3(U)
	\end{eqnarray*}
	so that $R^*(U) = \{$bad, dark, eatable, fruit, good, light, not, 
	thing$\}$ (see Figure \ref{F:reachable}). In particular, this means 
	that the word ``eatable" is $2$-reachable (but not $1$-reachable) from 
	$U$ and all words in $U$ are $0$-reachable from $U$. Moreover, we 
	observe that $U$ is not a grounding set of $G$ (``color", for example, 
	is unreachable). On the other hand, the set $U' = U \cup \{$or$\}$ is a 
	grounding set of $G$, so that $G$ is $5$-groundable.
\end{example}

\subsection{The Minimum Grounding Set Problem} \label{SS:mgsp}

Given a dictionary and its associated graph $G$, we are interested in finding 
minimum grounding sets of $G$. (Note that in general, there is more than one
grounding set of minimum cardinality.) This is related to a natural decision 
problem: we designate by $\kGS$ the problem of deciding whether $G$ is 
$k$-groundable. We show that $\kGS$ is closely related to the problem of 
finding minimum feedback vertex sets. First, we recall the definition of a 
feedback vertex set.

\begin{definition}
	Let $G = (V,E)$ be a directed graph and $U$ a subset of $V$. We say 
	that $U$ is a \emph{feedback vertex set} of $G$ if for every cycle $C$ 
	of $G$, we have $U \cap C \neq \emptyset$. In other words, $U$ covers 
	every cycle of $G$.
\end{definition}

The \emph{minimum feedback vertex set problem} is the problem of finding a 
feedback vertex set of $G$ of minimum cardinality. To show that feedback vertex 
sets and grounding sets are the same, we begin by stating two simple lemmas.

\begin{lemma}\label{L:cycle}
	Let $G = (V,E)$ be a directed graph, $C$ a cycle of $G$ and 
	$U \subseteq V$ a grounding set of $G$. Then $U \cap C \neq \emptyset$.
\end{lemma}

\begin{proof}
	By contradiction, assume that $U \cap C = \emptyset$ and, for all 
	$v \in C$, there exists an integer $k$ such that $v$ belongs to 
	$R^k(U)$. Let $\ell$ be the smallest index in the set \linebreak 
	$\{k \mid \exists u \in C \mbox{ such that } u \in R^k(U)\}$. Let $u$ 
	be a vertex in $C \cap R^{\ell}(U)$ and $w$ the predecessor of $u$ in 
	$C$. Since $U \cap C = \emptyset$, $k$ must be greater than $0$ and $w$ 
	a member of $R^{\ell-1}(U)$, contradicting the minimality of $\ell$.
\end{proof}

\begin{lemma}\label{L:acyclic}
	Every directed acyclic graph $G$ is $0$-groundable.
\end{lemma}

\begin{proof}
	We prove the statement by induction on $|V|$.
	\begin{basis}
		If $|V| = 1$, then $|E| = 0$, so that the only vertex $v$ of 
		$G$ satisfies $N_G^-(v) = \emptyset$. Hence $R(\emptyset) = V$.
	\end{basis}
	\begin{induction}
		Let $v$ be a vertex such that $\deg^+(v) = 0$. Such a vertex 
		exists since $G$ is acyclic. Moreover, let $G'$ be the 
		(acyclic) graph obtained from $G$ by removing vertex $v$ and 
		all its incident arcs. By the induction hypothesis, there 
		exists an integer $k$ such that 
		$R_{G'}^k(\emptyset) = V - \{v\}$. Therefore, 
		$V - \{v\} \subseteq R_G^k(\emptyset)$ so that 
		$R_G^{k+1}(\emptyset) = V$.
	\end{induction}
\end{proof}

The next theorem follows easily from Lemmas \ref{L:cycle} and \ref{L:acyclic}.

\begin{theorem} \label{T:equiv}
	Let $G = (V,E)$ be a directed graph and $U \subseteq V$. Then $U$ is a 
	grounding set of $G$ if and only if $U$ is a feedback vertex set of $G$.
\end{theorem}

\begin{proof}
	$(\Rightarrow)$ Let $C$ be a cycle of $G$. By Lemma \ref{L:cycle}, 
	$U \cap C \neq \emptyset$, so that $U$ is a minimum feedback vertex set 
	of $G$.
	$(\Leftarrow)$ Let $G'$ be the graph obtained from $G$ by removing $U$. 
	Then $G'$ is acyclic and $\emptyset$ is a grounding set of $G'$. 
	Therefore, $U \cup \emptyset = U$ is a grounding set of $G$.
\end{proof}

\begin{corollary}
	$\kGS$ is $\NPcomplete$.
\end{corollary}

\begin{proof}
	Denote by $\kFVS$ the problem of deciding whether a directed graph $G$ 
	admits a feedback vertex set of cardinality at most $k$. This problem 
	is known to be $\NPcomplete$ and has been widely studied 
	\cite{Karp72,Gare79}. It follows directly from Theorem \ref{T:equiv} 
	that $\kGS$ is $\NPcomplete$ as well since the problems are equivalent.
\end{proof}

The fact that problems $\kGS$ and $\kFVS$ are equivalent is not very 
surprising. Indeed, roughly speaking, the minimum grounding problem consists of 
finding a minimum set large enough to enable the reader to learn (reach) all 
the words of the dictionary. On the other hand, the minimum feedback vertex set 
problem consists of finding a minimum set large enough to break the circularity 
of the definitions in the dictionary. Hence, the problems are the same, even if 
they are stated differently.
	
Although the problem is $\NPcomplete$ in general, we show that there is a 
simple way of reducing the complexity of the problem by considering the 
strongly connected components.

\subsection{Decomposing the Problem} \label{SS:dividing}

Let $G = (V,E)$ be a directed graph and $G_1$, $G_2$, $\ldots$, $G_m$ the 
subgraphs induced by its strongly connected components, where $m \geq 1$. In 
particular, there are no cycles of $G$ containing vertices in different 
strongly connected components. Since the minimum grounding set problem is 
equivalent to the minimum feedback vertex set problem, this means that when 
seeking a minimum grounding set of $G$, we can restrict ourselves to seeking 
minimum grounding sets of $G_i$, for $i=1,2,\ldots,m$. More precisely, we have 
the following proposition.

\begin{proposition}
	Let $G = (V,E)$ be a directed graph with $m$ strongly connected 
	components, with $m \geq 1$, and let $G_i = (V_i,E_i)$ be the subgraph 
	induced by its $i$-th strongly connected component, where 
	$1 \leq i \leq m$. Moreover, let $U_i$ be a minimum grounding set of 
	$G_i$, for $i=1,2,\ldots,m$. Then $U = \bigcup_{i=1}^m U_i$ is a 
	minimum grounding set of $G$.
\end{proposition}

\begin{proof}
First, we show that $U$ is a grounding set of $G$. Let $C$ be a cycle of $G$. 
Then $C$ is completely contained in some strongly connected component of $G$, 
say $G_j$, where $1 \leq j \leq m$. But $U_j \subseteq U$ is a grounding set of 
$G_j$, therefore $U_j \cap C \neq \emptyset$ so that $U \cap C \neq \emptyset$. 
It remains to show that $U$ is a minimum grounding set of $G$. By 
contradiction, assume that there exists a grounding set $U'$ of $G$, with 
$|U'| < |U|$ and let $U'_i = U' \cap V_i$. Then there exists an index $j$, with 
$1 \leq j \leq m$, such that $|U'_j| < |U_j|$, contradicting the minimality of 
$|U_j|$.
\end{proof}

Note that this proposition may be very useful for graphs having many small 
strongly connected components. Indeed, by using Tarjan's Algorithm 
\cite{Tarj72}, the strongly connected components can be computed in linear 
time. We illustrate this reduction by an example.

\begin{figure}[ht]
	\includegraphics[width=\columnwidth]{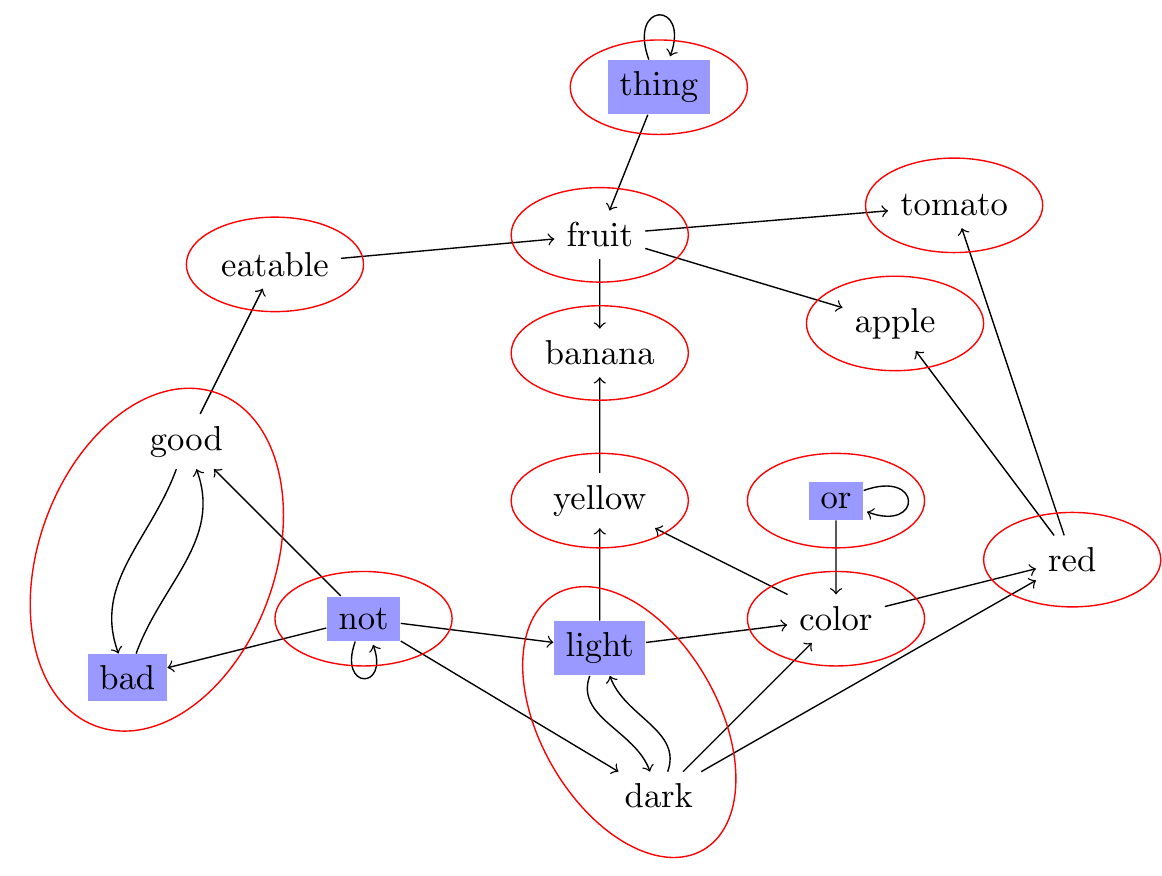}
	\caption{The strongly connected components and a minimum grounding set 
	                                                                of $G$}
	\label{F:strongly}
\end{figure}

\begin{example} \label{E:strongly}
Consider again the dictionary $D$ and the graph $G$ of Example 
\ref{E:dictionary}. The strongly connected components of $G$ are encircled in 
Figure \ref{F:strongly} and minimum grounding sets (represented by words in 
squares) for each of them are easily found. Thus the grounding number of $G$ is 
$5$.
\end{example}

\subsection{The Grounding Kernel}

In Example \ref{E:strongly}, we have seen that there exist some strongly 
connected components consisting of only one vertex without any loop. In 
particular, there exist vertices with no successor, i.e. vertices $v$ such that 
$N_G^+(v) = 0$. For instance, this is the case of the words ``apple", ``banana" 
and ``tomato", which are not used in any definition in the dictionary. Removing 
these three words, we notice that ``fruit", ``red" and ``yellow" are in the 
same situation and they can be removed as well. Pursuing the same idea, we can 
now remove the words ``color" and ``eatable". At this point, we cannot remove 
any further words. The set of remaining words is called the \emph{grounding 
kernel} of the graph $G$. More formally, we have the following definition..
\begin{definition}
	Let $D$ be a dictionary, $G = (V,E)$ its associated graph and 
	$G_1 = (V_1,E_1)$, $G_2 = (V_2,E_2)$, $\ldots$, $G_m = (V_m,E_m)$ the 
	subgraphs induced by the strongly	connected components of $G$, 
	where $m \geq 1$. Let $V'$ be the set of vertices $u$ such that $\{u\}$ 
	is a strongly connected component without any loop (i.e., $(u,u)$ is 
	not an arc of $G$). For any $u$, let $N^*(u)$ denote the set of 
	vertices $v$ such that $G$ contains a $uv$-path. Then the 
	\emph{grounding kernel} of $G$, denoted by $K_G$, is the set 
	$V - \{u \mid u \in V' \mbox{ and } N^*(u) \subseteq V' \}$.
\end{definition}

Clearly, every dictionary $D$ admits a grounding kernel, as shown by Algorithm 
\ref{A:groundingkernel}.
\begin{algorithm}[ht]
\caption{Computing the grounding kernel}
\label{A:groundingkernel}
\begin{algorithmic}[1]
	\Function{\GroundingKernel}{$G$}
		\State $G' \leftarrow G$
		\Repeat
			\State Let $W$ be the set of vertices of $G'$
			\State $U \leftarrow 
			            \{v \in W \mid N_{G'}^+(v) = \emptyset \}$
			\State $G' \leftarrow G'[W - U]$
		\Until{$U = \emptyset$}
		\State \Return $G'$
	\EndFunction
\end{algorithmic}
\end{algorithm}
Moreover, the grounding kernel is a grounding set of its associated graph $G$ 
and every minimum grounding set of $G$ is a subset of the grounding kernel. 
Therefore, in studying the symbol grounding problem in dictionaries, we can 
restrict ourselves to the grounding kernel of the graph $G$ corresponding to
$D$. This phenomenon is interesting because every dictionary contains many 
words that can be recursively removed without compromising the understanding of 
the other definitions. Formally, this property relates to the \emph{level} of a 
word: we will say of a word $w$ that it is of \emph{level} $k$ if it is 
$k$-reachable from $K_G$ but not $\ell$-reachable from $K_G$, for any 
$\ell < k$. In particular, level $0$ indicates that the word is part of the 
grounding kernel. A similar concept has been studied in \cite{Chan08}.

\begin{example}
	Continuing Example \ref{E:strongly} and from what we have seen so far, 
	it follows that the grounding kernel of $G$ is given by
	$$K_G = \{\mbox{bad, dark, good, light, not, or, thing}\}.$$
	Level $1$ words are ``color" and ``eatable", level $2$ words are 
	``fruit", ``red" and ``yellow", and level $3$ words are ``apple", 
	``banana" and ``tomato".
\end{example}

\section{Grounding Sets and the Mental Lexicon} \label{S:psycholinguistics}

In Section \ref{S:formulation}, we introduced all the necessary terminology to 
study the symbol grounding problem using graph theory and digital dictionaries. 
In this section, we explain how this model can be useful and on what 
assumptions it is based.

A dictionary is a formal symbol system. The preceding section showed how formal 
methods can be applied to this system in order to extract formal features. In 
cognitive science, this is the basis of \emph{computationalism} (or cognitivism 
or ``disembodied cognition" \cite{Pyly84}), according to which cognition, too, 
is a formal symbol system -- one that can be studied and explained 
independently of the hardware (or, insofar as it concerns humans, the wetware) 
on which it is implemented. However, pure computationalism is vulnerable to the 
problem of the grounding of symbols too \cite{Harn90}. Some of this can be 
remedied by the competing paradigm of embodied cognition 
\cite{Bars08,Glen02,Stee07}, which draws on dynamical (noncomputational) 
systems theory to ground cognition in sensorimotor experience. Although 
computationalism and symbol grounding provide the background context for our 
investigations and findings, the present paper does not favor any particular 
theory of mental representation of meaning.

A dictionary is a symbol system that relates words to words in such a way that 
the meanings of the definienda are conveyed via the definientes. The user is 
intended to arrive at an understanding of an unknown word through an 
understanding of its definition. What was formally demonstrated in Section 
\ref{S:formulation} agrees with common sense: although one can learn new word 
meanings from a dictionary, the entire dictionary cannot be learned in this way 
because of circular references in the definitions (\emph{cycles}, in graph 
theoretic terminology). Information -- \emph{nonverbal} information -- must 
come from outside the system to ground at least some of its symbols by some 
means other than just formal definition \cite{Cang01}. For humans, the two 
options are learned sensorimotor grounding and innate grounding. (Although the 
latter is no doubt important, our current focus is more on the former.)

The need for information from outside the dictionary is formalized in Section 
\ref{S:formulation}. Apart from confirming the need for such external 
grounding, we take a symmetric stance: In natural language, some word meanings 
--- especially highly abstract ones, such as those of mathematical or 
philosophical terms --- are not or cannot be acquired through direct 
sensorimotor grounding. They are acquired through the \emph{composition} of 
previously known words. The meaning of some of those words, or of the words in 
their respective definitions, must in turn have been grounded through direct 
sensorimotor experience. 

To state this in another way: Meaning is not just formal definitions all the 
way down; nor is it just sensorimotor experience all the way up. The two 
extreme poles of that continuum are \emph{sensorimotor induction} at one pole 
(trial and error experience with corrective feedback; observation, pointing, 
gestures, imitation, etc.), and \emph{symbolic instruction} (definitions, 
descriptions, explanation, verbal examples etc.) at the other pole. Being able 
to identify from their lexicological structure which words were acquired one 
way or the other would provide us with important clues about the cognitive 
processes underlying language and the mental representation of meaning.

To compare the word meanings acquired via sensorimotor induction with word 
meanings acquired via symbolic instruction (definitions), we first need access 
to the encoding of that knowledge. In this component of our research, our 
hypothesis is that the representational structure of word meanings in 
dictionaries shares some commonalities with the representational structure of 
word meanings in the human brain \cite{Hauk08}. We are thus trying to extract 
from dictionaries the grounding kernel (and eventually a minimum grounding set, 
which in general is a proper subset of this kernel), from which the rest of the 
dictionary can be reached through definitions alone. We hypothesize that this 
kernel, identified through formal structural analysis, will exhibit properties 
that are also reflected in the mental lexicon. In parallel ongoing studies, we 
are finding that the words in the grounding kernel are indeed (1) more frequent 
in oral and written usage, (2) more concrete, (3) more readily imageable, and 
(4) learned earlier or at a younger age. We also expect they will be (5) more 
universal (across dictionaries, languages and cultures) \cite{Chic08}. 

\section{Grounding Kernels in Natural Language Dictionaries} \label{S:kernels}

In earlier research \cite{Clar03}, we have been analyzing two special 
dictionaries: the Longman's Dictionary of Contemporary English (LDOCE) 
\cite{Proc78} and the Cambridge International Dictionary of English (CIDE) 
\cite{Proc95}. Both are officially described as being based upon a 
\emph{defining vocabulary}: a set of $2000$ words which are purportedly the 
only words used in all the definitions of the dictionary, including the 
definitions of the defining vocabulary itself. A closer analysis of this 
defining vocabulary, however, has revealed that it is not always faithful to 
these constraints: A significant number of words used in the definitions turn 
out not to be in the defining vocabulary. Hence it became evident 
that we would ourselves have to generate a grounding kernel (roughly equivalent 
to the defining vocabulary) from these dictionaries. 

The method presented in this paper makes it possible, given the graph structure 
of a dictionary, to extract a grounding kernel therefrom. Extracting this 
structure in turn confronts us with two further problems: \emph{morphology} and 
\emph{polysemy}. Neither of these problems has a definite algorithmic solution. 
Morphology can be treated through stemming and associated look-up lists for the 
simplest cases (\emph{i.e.}, was $\rightarrow$ to be, and children 
$\rightarrow$ child), but more elaborate or complicated cases would require 
syntactic analysis or, ultimately, human evaluation. Polysemy is usually 
treated through statistical analysis of the word context (as in Latent Semantic 
Analysis) \cite{Kint07} or human evaluation. Indeed, a good deal of background 
knowledge is necessary to analyse an entry such as: 
``\emph{dominant}: the fifth note of a musical scale of eight notes'' (the 
LDOCE notes 16 different meanings of \emph{scale} and 4 for \emph{dominant}, 
and in our example, none of these words are used with their most frequent 
meaning). 

Correct disambiguation of a dictionary is time-consuming work, as the most 
effective way to do it for now is through consensus among human evaluators. 
Fortunately, a fully disambiguated version of the WordNet database 
\cite{Fell98,Fell05} has just become available. We expect the grounding kernel 
of WordNet to be of greater interest than the defining vocabulary of either 
CIDE or LDOCE (or what we extract from them and disambiguate automatically, and 
imperfectly) for our analysis. 

\section{Future Work}

The main purpose of this paper was to introduce a formal approach to the symbol 
grounding problem based on the computational analysis of digital dictionaries. 
Ongoing and future work includes the following:

\emph{The minimum grounding set problem.} We have seen that the problem of 
finding a minimum grounding set is $\NPcomplete$ for general graphs. However, 
graphs associated with dictionaries have a very specific structure. We intend 
to describe a class of graphs including those specific graphs and to try to 
design a polynomial-time algorithm to solve the problem. Another approach is to 
design approximation algorithms, yielding a solution close to the optimal 
solution, with some known guarantee.

\emph{Grounding sets satisfying particular constraints.} Let $D$ be a 
dictionary, $G = (V,E)$ its associated graph, and $U \subseteq V$ any subset of 
vertices satisfying a given property $P$. We can use Algorithm 
\ref{A:reachableBFS} to test whether or not $U$ is a grounding set. In 
particular, it would be interesting to test different sets $U$ satisfying 
different cognitive constraints.

\emph{Relaxing the grounding conditions.} In this paper we imposed strong 
conditions on the learning of new words: One must know all the words of the 
definition fully in order to learn a new word from them. This is not realistic, 
because we all know one can often understand a definition without knowing every 
single word in it. Hence one way to relax these conditions would be to modify 
the learning rule so that one need only understand at least $r\%$ of the 
definition, where $r$ is some number between $0$ and $100$. Another variation 
would be to assign weights to words to take into account their morphosyntactic 
and semantic properties (rather than just treating them as an unordered list, 
as in the present analysis). Finally, we could consider ``quasi-grounding 
sets", whose associated reachable set consists of $r\%$ of the whole 
dictionary.

\emph{Disambiguation of definitional relations.} Analyzing real dictionaries 
raises, in its full generality, the problem of word and text disambiguation in 
free text; this is a very difficult problem. For example, if the word ``make"
appears in a definition, we do not know which of its many senses is intended 
--- nor even what its grammatical category is. To our knowledge, the only 
available dictionary that endeavors to provide fully disambiguated definitions 
is the just-released version of WordNet. On the other hand, dictionary 
definitions have a very specific grammatical structure, presumably simpler and 
more limited than the general case of free text. It might hence be feasible to 
develop automatic disambiguation algorithms specifically dedicated to the 
special case of dictionary definitions.

\emph{Concluding Remark}: Definition can reach the sense (sometimes), but only 
the senses can reach the referent.

\emph{Research funded by Canada Research Chair in Cognitive Sciences, SSHRC (S. 
Harnad) and NSERC (S. Harnad \& O. Marcotte)}


\begin{thebibliography}{}

\bibitem[\protect\citename{Barsalou}2008]{Bars08}
Barsalou, L.
\newblock (2008)
\newblock {\em Grounded Cognition}.
\newblock Annual Review of Psychology (in press).

\bibitem[\protect\citename{Bondy $\&$ Murty}1978]{Bond78}
Bondy, J.A. $\&$ U.S.R. Murty.
\newblock (1978)
\newblock {\em Graph theory with applications}.
\newblock Macmillan, New York.

\bibitem[\protect\citename{Cangelosi $\&$ Harnad}2001]{Cang01}
Cangelosi, A. $\&$ Harnad, S.
\newblock (2001)
\newblock {\em The Adaptive Advantage of Symbolic Theft Over Sensorimotor Toil: 
Grounding Language in Perceptual Categories}.
\newblock Evol. of Communication 4(1) 117-142.

\bibitem[\protect\citename{Changizi}2008]{Chan08}
Changizi, M.A.
\newblock (2008)
\newblock {\em Economically organized hierarchies in WordNet and the Oxford 
English Dictionary}.
\newblock Cognitive Systems Research (in press).

\bibitem[\protect\citename{Chicoisne et al.}2008]{Chic08}
Chicoisne G.,  A. Blondin-Mass\'e, O. Picard, S. Harnad
\newblock (2008)
\newblock {\em Grounding Abstract Word Definitions In Prior Concrete 
Experience}.
\newblock 6th Int. Conf. on the Mental Lexicon, Banff, Alberta.

\bibitem[\protect\citename{Clark}2003]{Clar03}
Clark G.
\newblock (2003)
\newblock {\em Recursion Through Dictionary Definition Space: Concrete Versus 
Abstract Words}.
\newblock (U. Southampton Tech Report).

\bibitem[\protect\citename{Fellbaum}1998]{Fell98}
Fellbaum, C.
\newblock (1998)
\newblock {\em WordNet: An electronic lexical database}.
\newblock Cambridge: MIT Press.

\bibitem[\protect\citename{Fellbaum}2005]{Fell05}
Fellbaum, C.
\newblock (2005)
\newblock {\em Theories of human semantic representation of the mental 
lexicon}.
\newblock In: Cruse, D. A. (Ed.), Handbook of Linguistics and Communication 
Science, Berlin, Germany: Walter de Gruyter, 1749-1758.

\bibitem[\protect\citename{Frege}1948]{Freg48}
Frege G.
\newblock (1948)
\newblock {\em Sense and Reference}.
\newblock The Philosophical Review 57 (3) 209-230.

\bibitem[\protect\citename{Garey $\&$ Johnson}1979]{Gare79}
Garey, M.R. $\&$ D.S. Johnson
\newblock (1979)
\newblock {\em Computers and Intractability: A Guide to the Theory of 
NP-completeness}.
\newblock W.H. Freeman, New York.

\bibitem[\protect\citename{Glenberg $\&$ Robertson}2002]{Glen02}
Glenberg A.M. $\&$ D.A. Robertson
\newblock (2002)
\newblock {\em Symbol Grounding and Meaning: A Comparison of High-Dimensional 
and Embodied Theories of Meaning}.
\newblock Journal of Memory and Language 43 (3) 379-401.

\bibitem[\protect\citename{Harnad}1990]{Harn90}
Harnad, S.
\newblock (1990)
\newblock {\em The Symbol Grounding Problem}.
\newblock Physica D 42:335-346.

\bibitem[\protect\citename{Harnad}2003]{Harn03}
Harnad, S.
\newblock (2003)
\newblock {\em Symbol-Grounding Problem}.
\newblock Encylopedia of Cognitive Science. Nature Publishing Group. Macmillan.

\bibitem[\protect\citename{Harnad}2005]{Harn05}
Harnad, S.
\newblock (2005)
\newblock {\em To Cognize is to Categorize: Cognition is Categorization}.
\newblock In Lefebvre, C. and Cohen, H. (Eds.), Handbook of Categorization. 
Elsevier.

\bibitem[\protect\citename{Hauk et al.}2008]{Hauk08}
Hauk, O., M.H. Davis, F. Kherif, F. Pulverm\"uller.
\newblock (2008)
\newblock {\em Imagery or meaning? Evidence for a semantic origin of 
category-specific brain activity in metabolic imaging}.
\newblock European Journal of Neuroscience 27 (7) 1856-1866. 

\bibitem[\protect\citename{Karp}1972]{Karp72}
Karp, R.M.
\newblock (1972)
\newblock {\em Reducibility among combinatorial problems}.
\newblock In: R.E. Miller, J.W. Thatcher (Eds.), Complexity of Computer 
Computations, Plenum Press, New York, 1972, pp. 85-103.

\bibitem[\protect\citename{Kintsch}2007]{Kint07}
Kintsch, W.
\newblock (2007)
\newblock {\em Meaning in Context}.
\newblock In T.K. Landauer, D.S. McNamara, S. Dennis $\&$ W. Kintsch (Eds.), 
Handbook of Latent Semantic Analysis. Erlbaum.

\bibitem[\protect\citename{Procter}1978]{Proc78}
Procter, P.
\newblock (1978)
\newblock {\em Longman Dictionary of Contemporary English}.
\newblock Longman Group Ltd., Essex, UK. 

\bibitem[\protect\citename{Procter}1995]{Proc95}
Procter, P.
\newblock (1995)
\newblock {\em Cambridge International Dictionary of English (CIDE)}.
\newblock Cambridge University Press. 

\bibitem[\protect\citename{Pylyshyn}1984]{Pyly84}
Pylyshyn, Z. W.
\newblock (1984)
\newblock {\em Computation and Cognition: Towards a Foundation for Cognitive
Science}.
\newblock Cambridge: MIT Press. 

\bibitem[\protect\citename{Ravasz $\&$ Barabasi}2003]{Rava03}
Ravasz, E. $\&$ Barabasi, A. L.
\newblock (2003)
\newblock {\em Hierarchical organization in complex networks}.
\newblock Physical Review E 67, 026112.

\bibitem[\protect\citename{Rosen}2007]{Rose07}
Rosen, K.H.
\newblock (2007)
\newblock {\em Discrete mathematics and its applications}, 6th ed.
\newblock McGraw-Hill.

\bibitem[\protect\citename{Steels}2007]{Stee07}
Steels, L.
\newblock (2007)
\newblock {\em The symbol grounding problem is solved, so what's next?}
\newblock In De Vega, M. and G. Glenberg and A. Graesser (Eds.), Symbols, 
embodiment and meaning. Academic Press, North Haven.

\bibitem[\protect\citename{Steyvers $\&$ Tenenbaum}2005]{Stey05}
Steyvers, M. $\&$ Tenenbaum J.B.
\newblock (2005)
\newblock {\em The large-scale structure of semantic networks: statistical 
analyses and a model of semantic growth}.
\newblock Cognitive Science, 29(1) 41-78. 

\bibitem[\protect\citename{Tarjan}1972]{Tarj72}
Tarjan, R.
\newblock (1972)
\newblock {\em Depth-first search and linear graph algorithms}.
\newblock SIAM Journal on Computing. 1 (2) 146-160. 

\end{thebibliography}
\end{document}